\documentclass[conference]{IEEEtran}
\usepackage{cite}
\usepackage{tabularx}
\usepackage[utf8]{inputenc}
\usepackage{amsmath,amssymb,amsfonts}
\usepackage{algorithm, algpseudocode} 
\DeclareUnicodeCharacter{2060}{}
\DeclareUnicodeCharacter{2212}{} 
\usepackage{graphicx}
\usepackage{subcaption} 
\usepackage{float} 
\usepackage{textcomp}
\usepackage{xcolor}
\usepackage{hyperref}
\usepackage{threeparttable} 

\def\BibTeX{{\rm B\kern-.05em{\sc i\kern-.025em b}\kern-.08em
    T\kern-.1667em\lower.7ex\hbox{E}\kern-.125emX}}

\begin{document}

\title{SOAR: Advancements in Small Body Object Detection for Aerial Imagery Using State Space Models and Programmable Gradients
}

\author{\IEEEauthorblockN{Tushar Verma}
\IEEEauthorblockA{\textit{Netaji Subhas University of Technology}\\
0009-0006-0299-9702}
\and
\IEEEauthorblockN{Yash Bhartari}
\IEEEauthorblockA{\textit{Netaji Subhas University of Technology}\\
0009-0008-9465-5572}
\and
\IEEEauthorblockN{Rishi Jarwal}
\IEEEauthorblockA{\textit{Netaji Subhas University of Technology}\\
0009-0008-4179-5045}
\and
\IEEEauthorblockN{Suraj Singh}
\IEEEauthorblockA{\textit{Netaji Subhas University of Technology}\\
0009-0008-8696-0876}
\and
\IEEEauthorblockN{Shubhkarman Singh}
\IEEEauthorblockA{\textit{Netaji Subhas University of Technology}\\
0009-0000-5602-5432}
\and
\IEEEauthorblockN{Jyotsna Singh}
\IEEEauthorblockA{\textit{Netaji Subhas University of Technology}\\
}
}
\maketitle

\begin{abstract}
Small object detection in aerial imagery presents significant challenges in computer vision due to the minimal data inherent in small-sized objects and their propensity to be obscured by larger objects and background noise. Traditional methods using transformer-based models often face limitations stemming from the lack of specialized databases, which adversely affect their performance with objects of varying orientations and scales. This underscores the need for more adaptable, lightweight models. In response, this paper introduces two innovative approaches that significantly enhance detection and segmentation capabilities for small aerial objects. Firstly, we explore the use of the SAHI framework on the newly introduced lightweight YOLO v9 architecture, which utilizes Programmable Gradient Information (PGI) to reduce the substantial information loss typically encountered in sequential feature extraction processes. The paper employs the Vision Mamba model, which incorporates position embeddings to facilitate precise location-aware visual understanding, combined with a novel bidirectional State Space Model (SSM) for effective visual context modeling. This State Space Model adeptly harnesses the linear complexity of CNNs and the global receptive field of Transformers, making it particularly effective in remote sensing image classification. Our experimental results demonstrate substantial improvements in detection accuracy and processing efficiency, validating the applicability of these approaches for real-time small object detection across diverse aerial scenarios. This paper also discusses how these methodologies could serve as foundational models for future advancements in aerial object recognition technologies. The source code will be made accessible here 
\href{https://github.com/yash2629/S.O.A.R}{https://github.com/yash2629/S.O.A.R}
\end{abstract}


\section{Introduction}A fundamental component of many applications, ranging from video surveillance to intelligent traffic management and digital city infrastructure, is the detection and tracking of objects in diverse visual media, object detection identifies a target object inside a single frame or image. These qualities are critical for many computer vision applications. The biggest challenge seems to be Small Object detection. An object is considered little if it appears tiny in the video frame, usually because it was monitored from a distance. Their tiny size sometimes leads to them being confused for noise, which has a detrimental effect on tracking accuracy. The small objects can be defined as MS-COCO \cite{lin2014microsoft} metric evaluation, small objects are defined as having an area of 32 × 32 pixels or less, which is a commonly used threshold for datasets including common objects. There have been some models that have been developed to help in the particular problem and continuous improvement is being done on them too.
\par
While the results from currently available oriented aerial detectors are encouraging, they primarily concentrate on orientation modeling and pay little attention to object size. Identifying the area contained by bounding boxes using object detection is a useful method for comprehending things in an image by explaining what these objects are and where they are. Using a rectangular bounding box without an angle orientation also known as a horizontal bounding box (HBB) is the standard procedure. The model must be able to locate an object with accuracy and recognize its class to contain it inside an HBB.
\par
However, this method is particularly unsuccessful at detecting oriented aerial objects; more noise and background will be enclosed, which can cause misdetection; objects cannot be properly localized. Consequently, an orientated bounding box (OBB) producing object detector was introduced. Techniques already in use aided in the creation of efficient OBB detectors that can precisely encompass orientated objects. Refinement features \cite{han2021align, yang2021r3det, yang2019scrdet}, proposal extraction \cite{ding2019learning, han2021redet, li2022oriented, ma2018arbitrary}, orientation alignment \cite{li2022oriented, xu2020gliding}, and regression loss design \cite{chen2020piou, yang2021rethinking, yang2021learning} are some of the approaches that fall under this category.
\par
Despite the improved performance, the work on small object detection continued because of the combination of multi-scale prediction, shallow and deep networks, and more loss functions for large-scale object recognition while small objects were disregarded. Keeping all that in mind work on some new models with multi-scale feature fusion was encouraged. Adapting towards the same path this model takes inspiration from these developments and has to concentrate more on the expression of physical information about small items through multi-scale detection to achieve accurate small object detection. To anticipate small objects, the algorithm will be needed to combine deeper backbone networks and additional scales. The concept of a feature pyramid can be applied to more sensibly express the physical information of the shallow network and the semantic properties of the deep network. But previously it has come to notice that this can lead to an issue such as the backbone network getting deeper, the network parameter growing as well, and the quantity of the computation. Further efforts can surely be made to enhance the backbone network and achieve faster detection rates while maintaining a high level of detection accuracy.  
\par
While the use of deep learning architectures has led to the development of very accurate techniques like RetinaNet \cite{lin2017focal}, VarifocalNet \cite{zhang2021varifocalnet}, Cascade R-CNN \cite{cai2018cascade}, and Faster R-CNN \cite{ren2016faster}, these techniques are not without their variations. All of these new detectors are tested and trained on popular datasets such as MS COCO \cite{lin2014microsoft}, Pascal VOC12 \cite{everingham2010pascal}, and ImageNet \cite{deng2009large}. These datasets mostly consist of low-resolution photos (640 x 480) that feature huge objects with extensive pixel coverage, typically encompassing 60\% of the image height. Long-range object detection that satisfies the Detection, Observation, Recognition, and Identification (DORI) \cite{cagatay2022slicing} requirements has been made possible by those developments. It was suggested to slice to maintain higher memory utilization while aiding inference and fine-tuning for small item detection on high-resolution photos. To create the detection, use anchor boxes and categorize every point on the feature pyramid \cite{lin2017feature} as either background or foreground. Then forecast the distances between the foreground point and the four corners of the ground-truth bounding box directly.
\par
Talking about more on Small Object detection the YOLO series is the most popular real-time object detector, it is currently the standard in real-time object detection. YOLOv7 has been utilized in the past for small object recognition and has shown to be successful in a range of computer vision tasks and settings. The newly introduced YOLOv9 is believed to be the best real-time object detector of the new generation due to the aforementioned innovative technique. This, when combined with SAHI, can significantly enhance the new model's capability over its predecessors.
\par
It has been discovered that the generic slicing supports an inference pipeline that can be applied on top of any object detector in use thus helping with its fine-tuning. In this manner, slicing-assisted inference improved the small object detection performance of any object detector that is currently on the market without the need for fine-tuning and there is no need for pretraining when integrating the slicing-assisted hyper inference scheme into any object detection inference pipeline. Furthermore, optimizing the pre-trained models resulted in an extra performance benefit.\\
\\

\textbf{Review of Contributions:}
\begin{itemize}
    \item YOLOv9, a novel aerial imagery platform, is deployed on DOTA. It improves upon prior works by combining sliced-aided hyper inferencing pipeline adapters with pretraining regimes.
    \item A new framework for dynamic small-body object detection and experimental validation using SOAR over Vision Mamba architecture.
    \item Proposal of a novel framework towards fusion of Programmable gradient information with state space model representation for effective visual and computer vision task settings.
\end{itemize}

\section{Related Works}

In recent years, object detection research has witnessed significant advancements propelled by innovative methodologies and algorithms tailored to address specific challenges. This section presents a review of relevant literature focusing on enhancing object detection performance, particularly in scenarios involving small or distant objects.
\par
The region-based convolutional neural network (RCNN) was one of the first successful deep-learning approaches to detect objects. The performance was achieved through two insights. The first was to apply high-capacity convolutional neural networks to bottom-up region proposals to localize and segment objects. The second was a paradigm for training large CNNs when labeled training data is scarce. Also, earlier work combined handcrafted characteristics with deep learning-based features to enhance object detection, from the aegis within the YOLOv3 architecture. By leveraging hierarchical representations of Convolutional Neural Networks (CNNs) and domain-specific information stored in handmade features, the method overcomes the limitations of previous techniques. Through careful selection of convolutional layers for feature injection and refining feature combinations, substantial improvements in mean Average Precision (mAP) are observed compared to YOLOv3 on benchmark datasets such as PASCAL-VOC and MS-COCO. The study emphasizes the potential of feature fusion techniques in object detection, demonstrating considerable gains in detection robustness and accuracy.

\par
Another notable contribution comes from the MSFYOLO algorithm. This algorithm enhances small object detection by integrating a side path for feature re-fusion within the FPN \cite{lin2017feature} architecture, based on the PANET \cite{zheng2020distance}  framework. Through thoughtful loss function design and extensive testing, MSFYOLO outperforms existing methods such as YOLOv5 and RetinaNet, especially in challenging conditions, exhibiting robustness and high Frames Per Second (FPS)\cite{lin2017feature}. The algorithm's effectiveness positions it for real-world applications where both efficiency and accuracy are critical, spanning fields like industrial inspection, autonomous driving, surveillance, and medical imaging.
\par
Addressing the challenges of detecting small and distant objects in surveillance imagery, the SAHI framework introduces a slicing-aided hyper-inference approach. This method seamlessly integrates into existing object detection systems without pretraining, enhancing the detectability of small objects by slicing input images into overlapping patches during both fine-tuning and inference. Experimental evaluations demonstrate significant Average Precision (AP) improvements, particularly for small objects, across multiple datasets and detectors. SAHI offers a practical solution for small object detection in surveillance applications, with potential implications across various domains.
\par
Lastly, the study on Multiscale Faster-RCNN \cite{ren2016faster} addresses challenges in small object detection in machine vision. Leveraging multiscale feature extraction, the proposed approach demonstrates superior small object detection compared to Faster-RCNN \cite{ren2016faster}, validated through experimental evaluation and real-world scenarios. The study underscores the significance of tackling small object detection challenges and suggests avenues for future exploration, such as employing advanced techniques like Generative Adversarial Networks (GANs).
\par
Collectively, these studies represent significant strides in the field of object detection, offering diverse methodologies and algorithms tailored to enhance detection performance, particularly in scenarios involving small or distant objects.

\begin{figure*}[t]
    \centering
    \includegraphics[width=\linewidth]{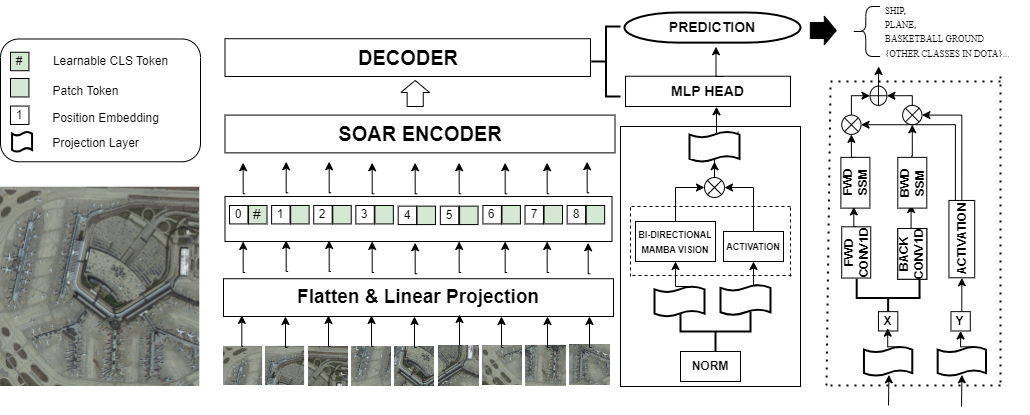}
    \caption{The architecture of the SOAR framework}
    \label{fig:architecture}
\end{figure*}

\section{Methodology}

\subsection{Preliminaries}
The State Space Model (SSM) is designed to characterize one-dimensional functions or sequences, where $u(t) \in \mathbb{R}$ maps to $v(t) \in \mathbb{R}$ via a concealed state $z(t) \in \mathbb{R}^N$. Employing $E$ as the evolution parameter and $F$ and $G$ as the projection parameters, the system operates with the following equations:

\begin{equation}
z(t) = Ez(t) + Fu(t),\\
\quad v(t) = Gz(t) \end{equation}

To discretize the continuous parameters, a timescale parameter $\Delta$ is employed, a technique commonly referred to as zero-order hold, expressed as follows:

\begin{equation}
\overline{E} = \exp(\Delta E),
\quad \overline{F} = (\Delta E)^{-1}(\exp(\Delta E) - I) \cdot \Delta F    
\end{equation}

The discretized version of the original equations is represented as:

\begin{equation}
z_t = E z_{t-1} + F u_t, \quad v_t = G z_t    
\end{equation}

A step of $\Delta$ is utilized in these equations.

Finally, the model calculates the output via global convolution:
\begin{equation}
K = (G \overline{F}, G \overline{E F}, \ldots, G \overline{E^{M-1} F}), \\
\quad v = u * K,
\end{equation}

where $M$ signifies the length of the sequence and $K \in \mathbb{R}^M$ denotes a structured convolution kernel.

\subsection{⁠Programmable Gradients Information}
The YOLOv8 model faced problems with small objects in images. Minimal pixel objects were difficult to accurately represent because the model's receptive field might not be able to capture enough information. According to the principle of information bottleneck, the data causes information loss when going through transformations as shown in Equation (5).

\begin{equation}
I(X, X) \geq I(X, f(X)) \geq I(X, g_{\phi}(f_{\theta}(X)))    
\end{equation}

Where $X$ indicates the data, $I$ indicates mutual information, $f$ and $g$ are transformation equations, and $\theta$ and $\phi$ are parameters of $f$ and $g$ respectively.

The following information is used to tell that as the number of network layers increases there will be a loss of original data. To decrease this information loss, reversible functions are used as shown in Equation (6):
\begin{equation}
I(X, X) = I(X, r_{\psi}(X)) = I(X, v_{\zeta}(r_{\psi}(X)))    
\end{equation}
where $\psi$ and $\zeta$ are parameters of $r$ and $v$, and for the application of the above method on lightweight models, the concept of information bottleneck is used. The formula for the same is:
\begin{equation}
I(X, X) \geq I(Y, X) \geq I(Y, f_{\theta}(X)) \geq \ldots \geq I(Y, \hat{Y})
\end{equation}
Where $I(Y, X)$ will only occupy a very small part of $I(X, X)$. To counter all the aforementioned problems, a programmable gradient information method is used, which uses an auxiliary reversible branch and multi-level auxiliary information.

The YOLOv9 \cite{wang2024yolov9} model also uses a Generalised ELAN network made from combining ELAN \cite{wang2022designing} and CSPNet \cite{wang2020cspnet}. For improved learning of small objects in the usually large datasets, hyper-inferencing is used. It is applied on top of the YOLOv9 model for object detection, providing a generic slicing-aided inference and fine-tuning pipeline. Sliced Aided Hyper Inferencing \cite{akyon2022slicing} divides the images into overlapping patches which result in a larger pixel area for small objects. It has a 2-part process combining slice-aided fine-tuning and slice-aided hyperinferencing. During the fine-tuning process, the data is enlarged by extracting patches and resizing them into larger images. Each image $I_{F1}, I_{F2}, \ldots, I_{Fj}$ is sliced into overlapping patches $P_{F1}, P_{F2}, \ldots, P_{Fk}$ with dimensions $M$ and $N$ selected within predefined ranges $[M_{\text{min}}, M_{\text{max}}]$ and $[N_{\text{min}}, N_{\text{max}}]$, which are treated as hyper-parameters. Then the patches are resized taking care of the aspect ratio, which results in larger object sizes as compared to that of the original image, while during the process of inferencing, the image is divided into smaller patches and then object detection forward pass is applied to each of the independent patches. Finally, the overlapping prediction results are merged into the original image.

\subsection{Overall architecture}

An overview of the proposed SOAR encoder is shown in Fig.\ref{fig:architecture}. To streamline the processing of visual data, we start by transforming the 2D image $I \in \mathbb{R}^{H \times W \times C}$ into flattened 2D patches $P \in \mathbb{R}^{J \times (S^2 \cdot C)}$, where $(H, W)$ denotes the dimensions of the input image, $C$ represents the number of channels, and $S$ denotes the size of the image patches. Subsequently, a linear projection of $P$ onto a vector with dimensionality $D$ is conducted, integrating position embeddings $E_{\text{pos}} \in \mathbb{R}^{(J+1) \times D}$ using the equation:

\begin{equation}
    V_0 = [V_{\text{cls}};  V^1_{P} \cdot Z; V^2_{P} \cdot Z ; \ldots; V^J_{P}\cdot Z] + E_{\text{pos}}
\end{equation}

Here, $V_{\text{cls}}$ denotes the class token, $V^j_{P}$ signifies the $j$-th patch of $I$, $Z \in \mathbb{R}^{(S^2 \cdot C) \times D}$ represents the learnable projection matrix, and $J$ indicates the total number of patches. This procedure facilitates the preparation of image data for subsequent analysis, embedding contextual information through position embeddings. Inspired by Vision Mamba \cite{zhu2024vision} and ViT \cite{dosovitskiy2020image}, we then forward the token sequence $V_{l-1}$ to the $l$-th layer of the SOAR encoder to produce $V_{l}^o$. In the SOAR encoder, the input token sequence $V_{l-1}$ undergoes normalization by the normalization layer. Subsequently, the normalized sequence is linearly projected to $x$ and $z$ with a dimension size $E$. Following this, $x$ is processed in both forward and backward directions as shown in (1). For each direction, a 1-D convolution is applied to $x$, yielding $x'_o$. Subsequently, $x'_o$ is linearly projected to $F_o$, $G_o$, and $\Delta_o$ as in (2). This $\Delta_o$ is subsequently used to obtain $F_o$, $E_o$ post which $V_\text{forward}$ and $V_\text{backward}$ are processed by the State Space Model.

\begin{equation}
    Q = \text{Norm}(V_L^o) 
\end{equation}
\begin{equation}
    T = \text{MLP}(Q)
\end{equation}


The introduction of a programmable gradient information layer fused on top of the decoder block consisting of an MLP head is further proposed and exploited as a promising future work toward the fusion of these novel architectures.

\section{Experimentation and Results}
In this section, we evaluate our neuro-graphic retrieval agent framework through quantitative and qualitative experiments, aiming to assess its performance and versatility across diverse and complex tasks.
\subsection{Experimental Setup}
We implemented our approach in PyTorch. We used 2 machines with a 24GB GPU server each. We used the Distributed Data Parallelism (DDP) technique for multi-machine training since the model was too large to fit on a single GPU which was available to us. 

\subsection{Datatset}
The datasets used for small object detection and tracking are essential for evaluating and benchmarking the performance of various algorithms in this field and the research conducted herein relies on the comprehensive and diverse dataset provided by DOTA \cite{ding2021object} specifically DOTAv1.5. DOTA is a sizable dataset used for aerial image object detection. It can be applied to the development and assessment of aerial picture object detectors. The photos are gathered from various platforms and sensors. Every image has a pixel size ranging from 800 x 800 to 20,000 x 20,000 pixels. The China Centre for Resources Satellite Data and Application, Google Earth, GF-2 and JL-1 satellites, and aerial photos from CycloMedia B.V. are the sources of the DOTA-v1.5 photographs. DOTA is made up of both grayscale and RGB images. The grayscale images are from the panchromatic band of GF-2 and JL-1 satellite photographs, while the RGB images are from Google Earth and CycloMedia. Every image is kept in a 'png' format. Planes, ships, storage tanks, baseball diamonds, tennis courts, basketball courts, ground track fields, harbors, bridges, big vehicles, small vehicles, helicopters, roundabouts, soccer ball fields, swimming pools, and container cranes are the objects types in DOTA-v1.5. There were a total of 1869 images in total in the dataset which included 1410 train sets, 438 valid sets, and 21 test sets. The preprocessing Grayscale was applied. We converted labels from dota format to coco format using dota dev kit.

\subsection{Implementation Details}
\begin{figure}[th]
    \centering
    \begin{subfigure}[b]{0.5\linewidth}
        \includegraphics[width=\linewidth]{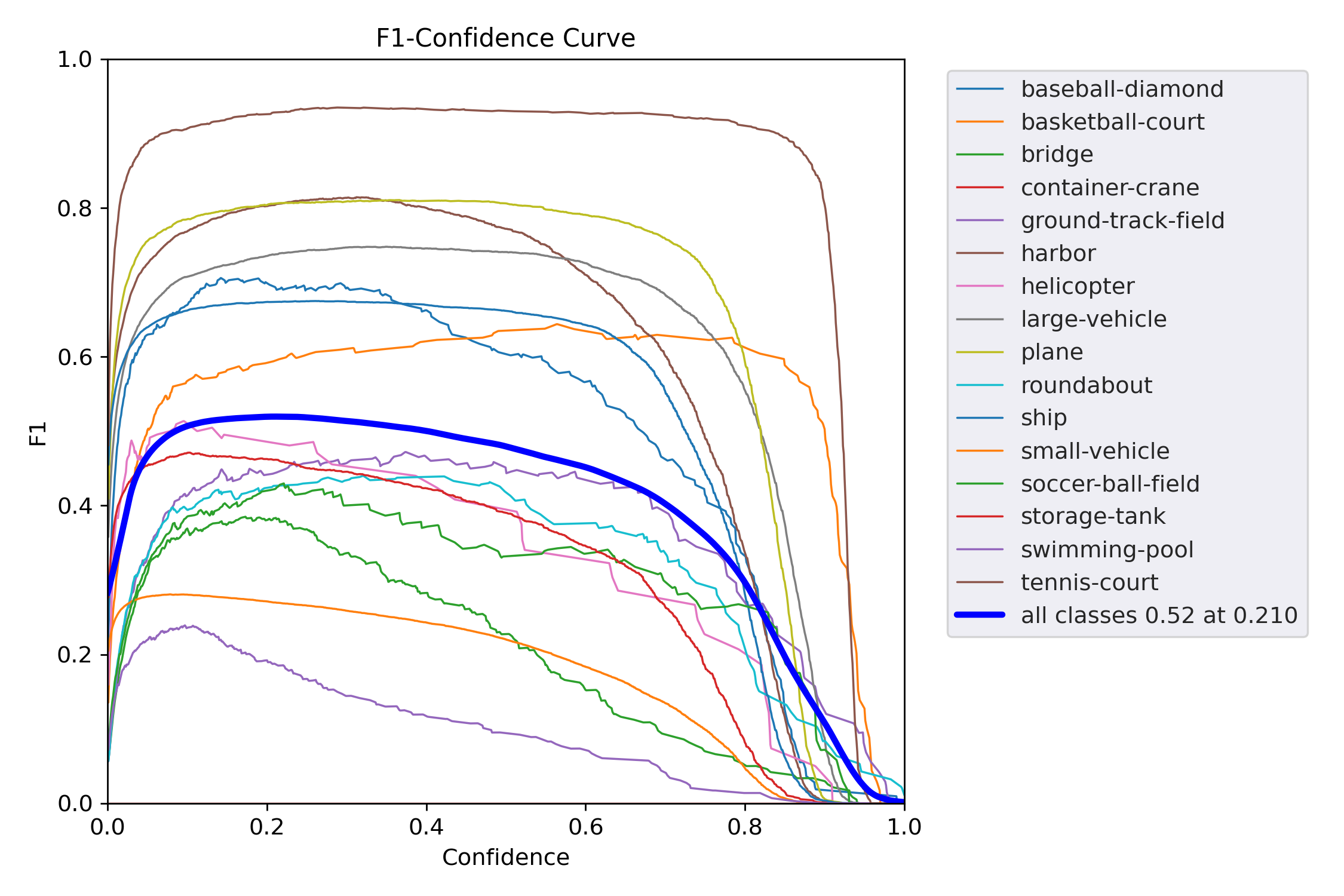 }
        \label{fig:f1}
    \end{subfigure}%
    \begin{subfigure}[b]{0.5\linewidth}
        \includegraphics[width=\linewidth]{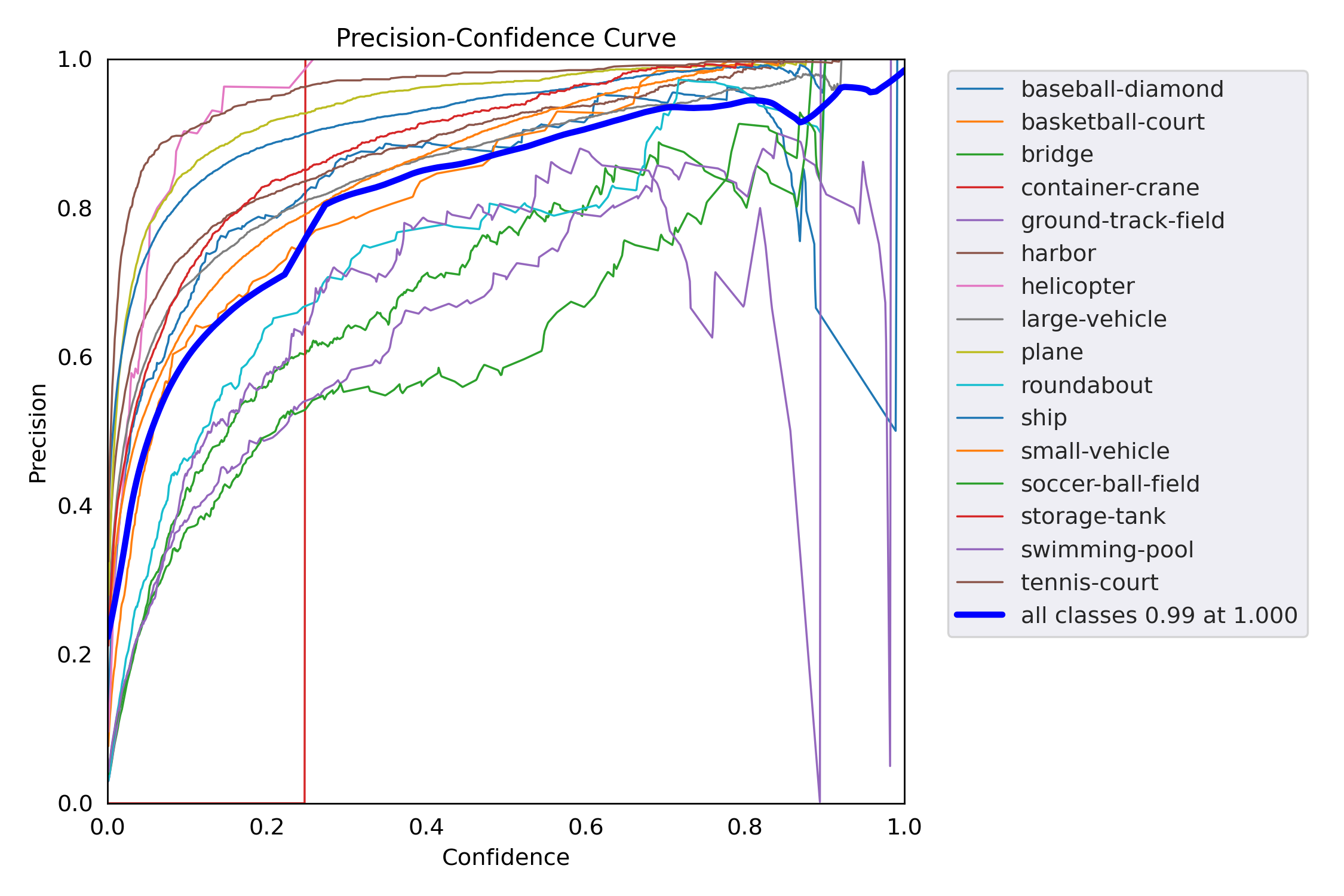 }
        \label{fig:f2}
    \end{subfigure}
    \begin{subfigure}[b]{0.5\linewidth}
        \includegraphics[width=\linewidth]{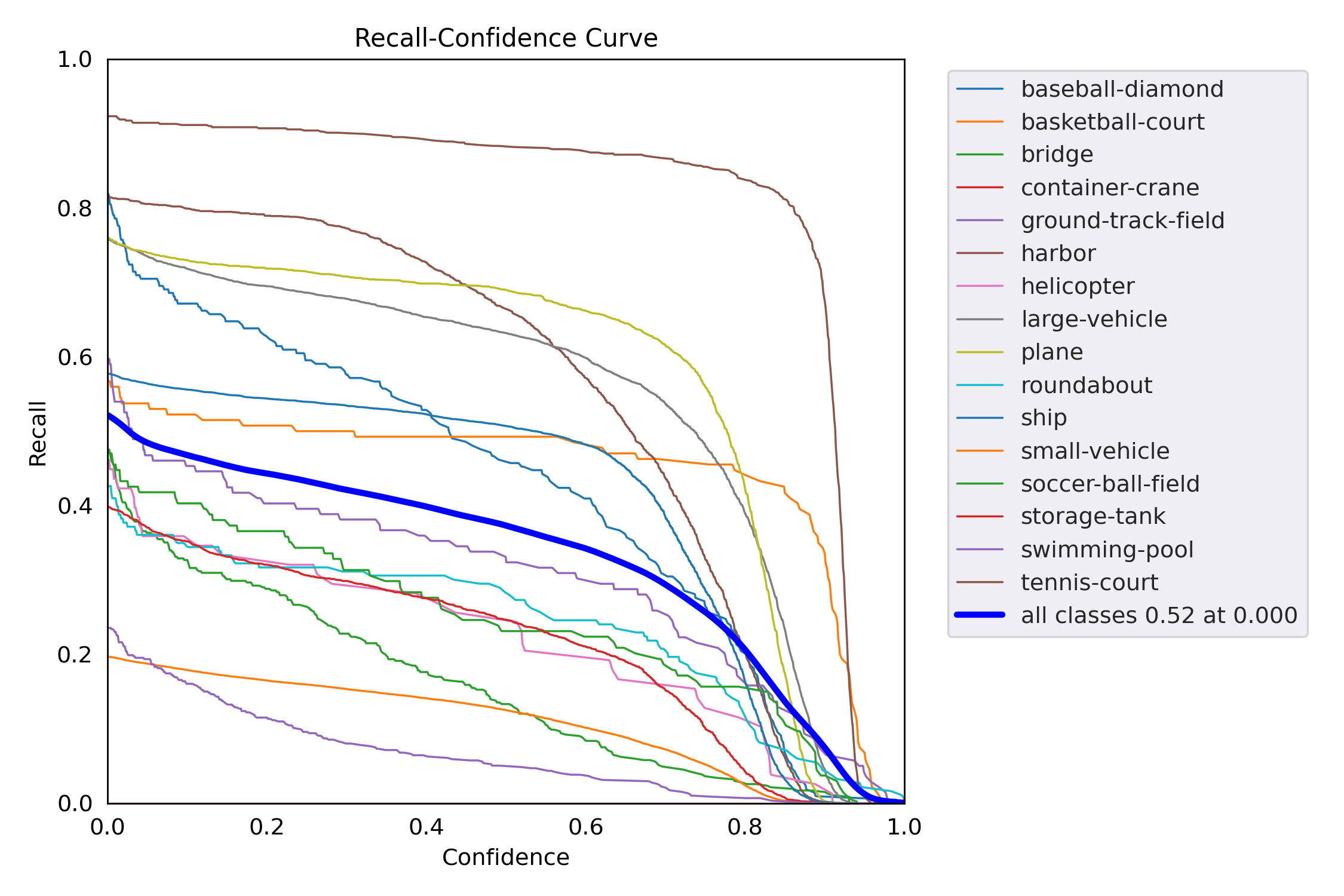 }
        \label{fig:f3}
    \end{subfigure}%
    \begin{subfigure}[b]{0.5\linewidth}
        \includegraphics[width=\linewidth]{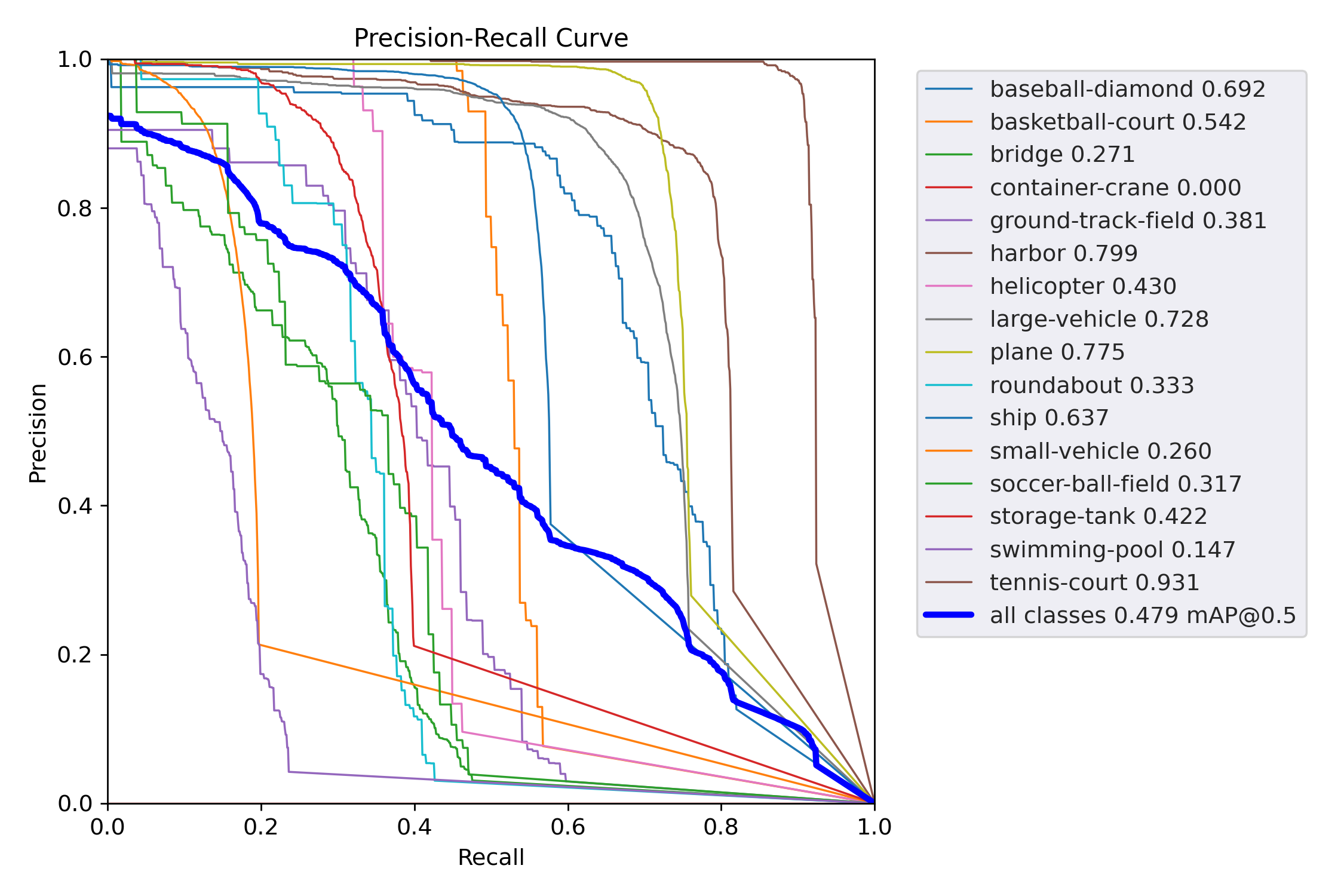 }
        \label{fig:f4}
    \end{subfigure}
    \caption{F1, Precision, Recall, and PR curves}
    \label{fig:combined-figures-fpr}
\end{figure}

\begin{figure}[htbp]
    \centering
    \begin{subfigure}{\linewidth}
        \centering
        \includegraphics[width=\linewidth]{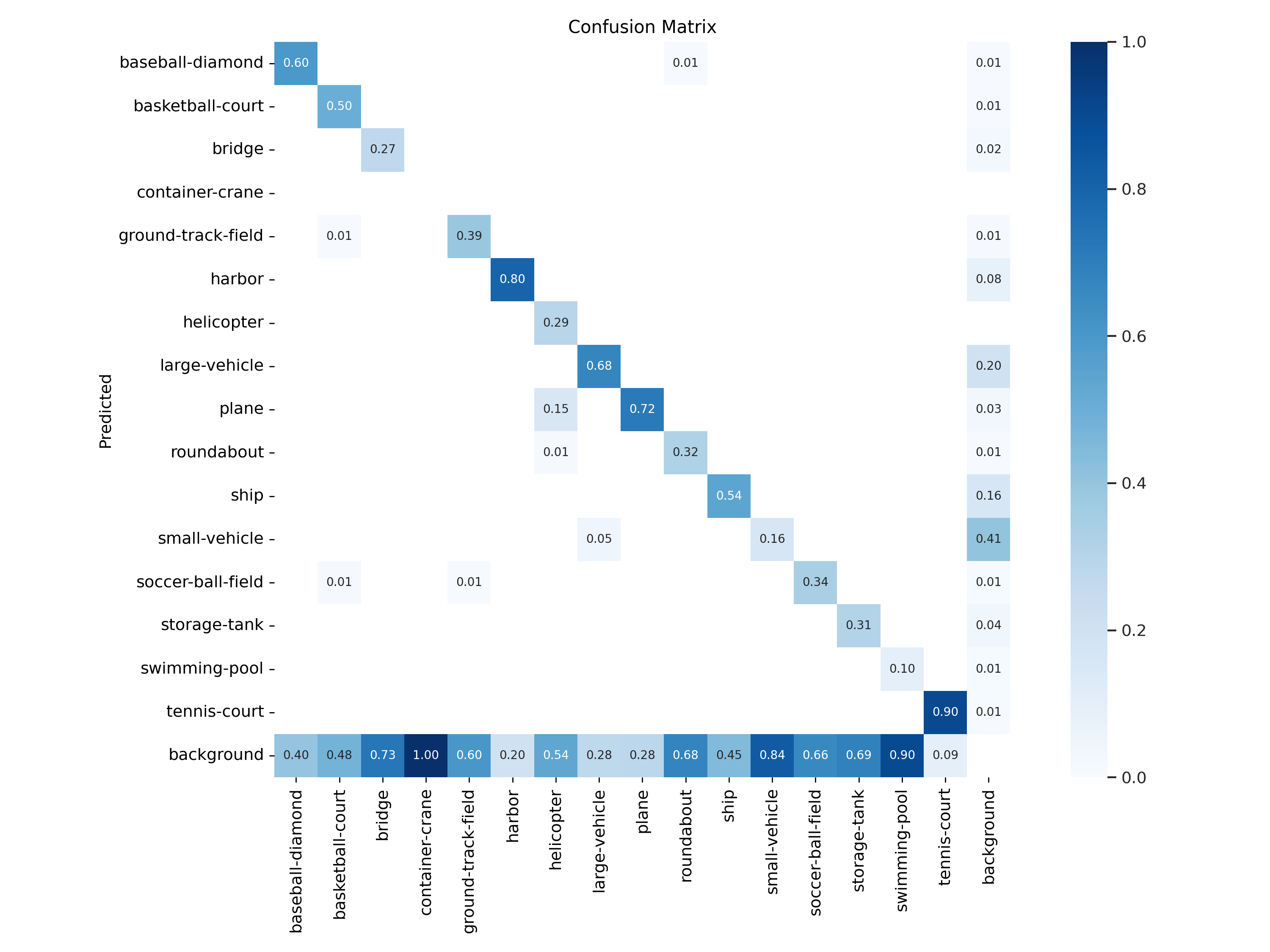}
        \caption{Confusion Matrix}
        \label{fig:cm}
    \end{subfigure}

    \medskip 

    \begin{subfigure}{\linewidth}
        \centering
        \includegraphics[width=\linewidth]{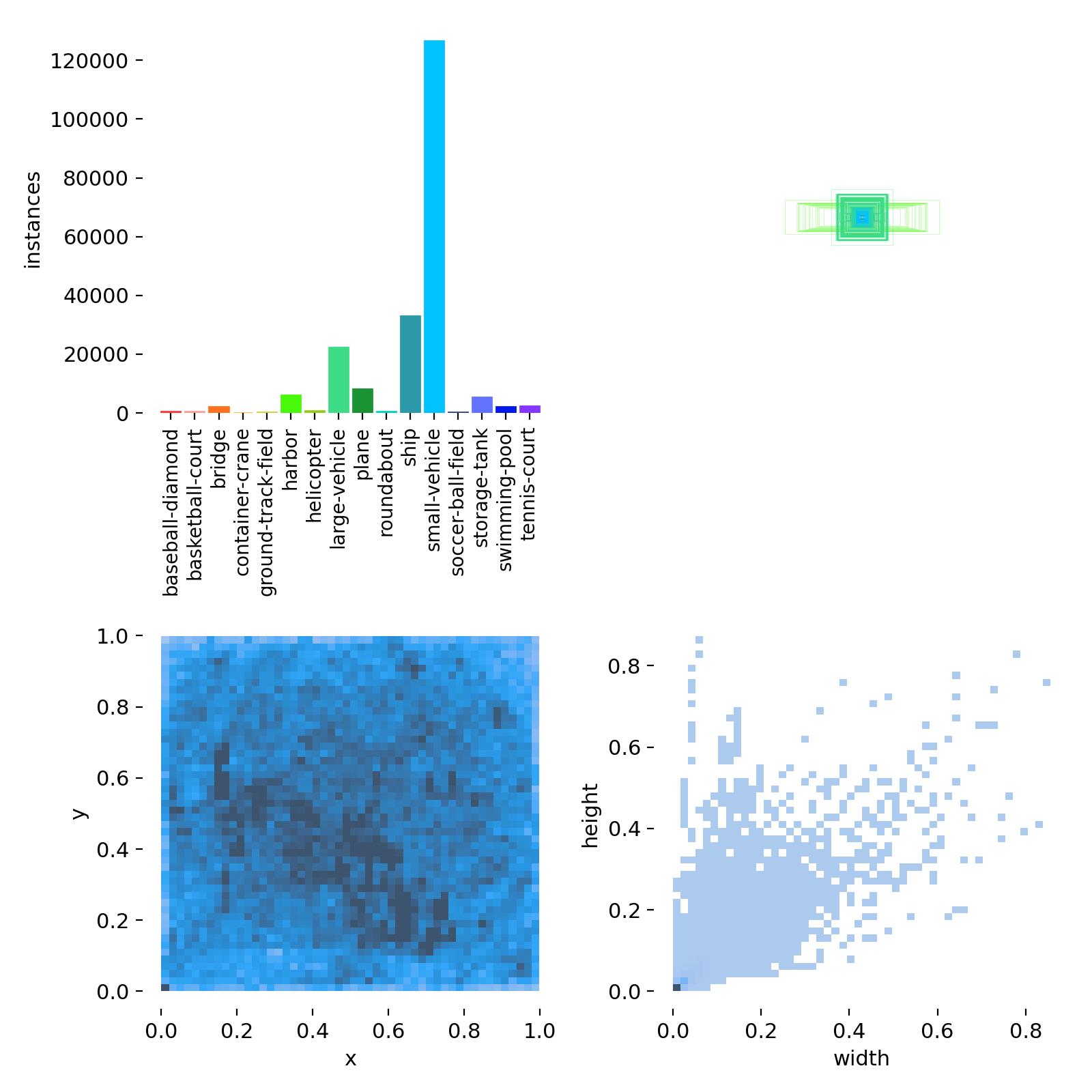}
        \label{fig:labels}
    \end{subfigure}
    \medskip 
    \begin{subfigure}{\linewidth}
        \centering
        \includegraphics[width=\linewidth]{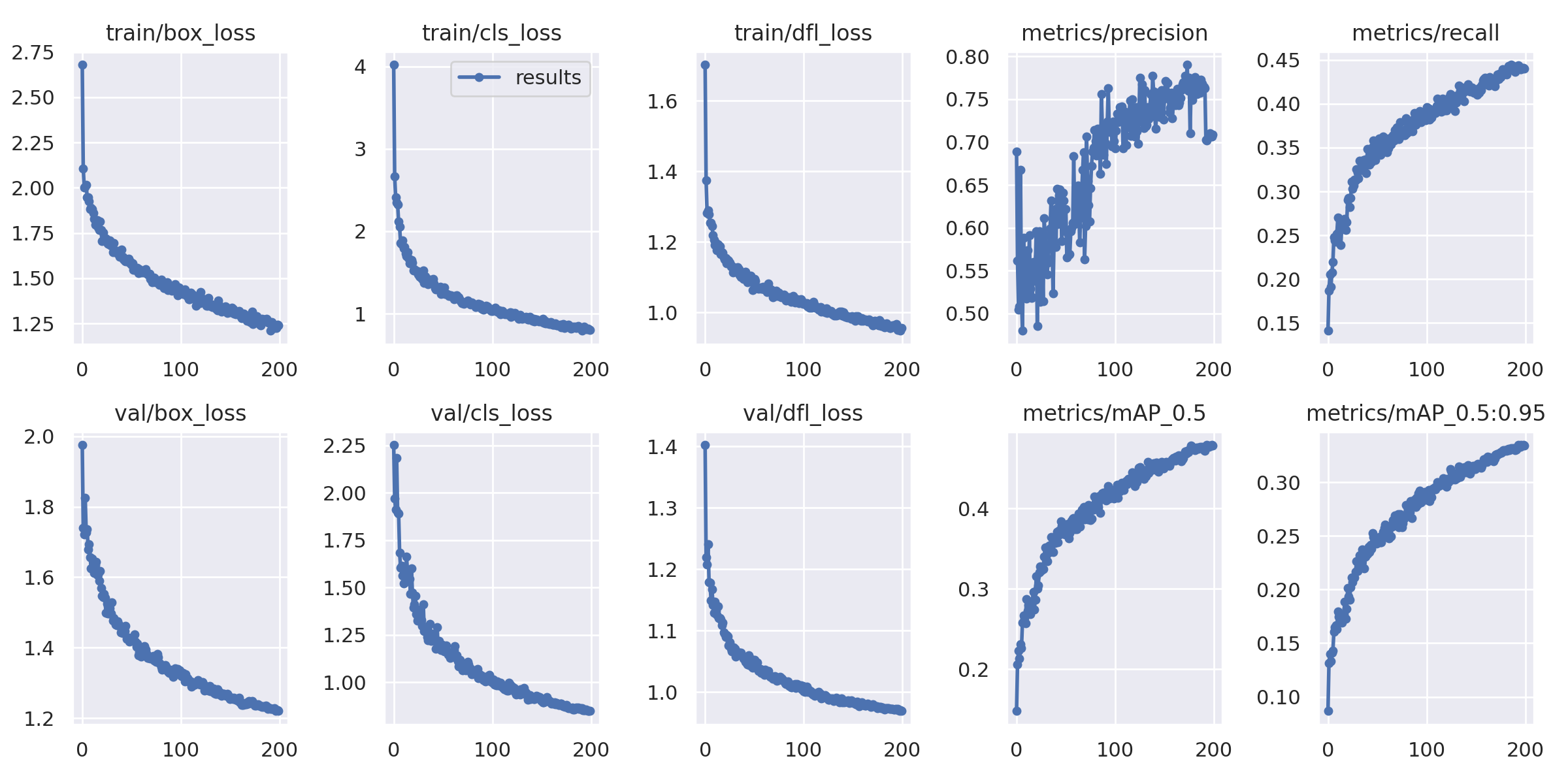}
        \caption{Train v/s Loss curves}
        \label{fig:train}
    \end{subfigure}
    \medskip 
    \medskip 
\medskip 

    \caption{Results and Analysis}
    \label{fig:combined-results}
\end{figure}

In our paper, we employ a fixed input image size of 224 × 224 and implement data augmentation techniques including random cropping, flipping, photometric distortion, mixup, cutMix, etc. Images are processed into sequential data through a two-dimensional convolution with a kernel size of 16 (k = 16) and a stride of 8 (s = 8). Position encodings are represented by randomly initialized learnable parameters. For supervised training, we employ the cross-entropy loss function and utilize the AdamW optimizer with an initial learning rate of 5e − 4 and a weight decay of 0.05. The learning rate is decayed using a cosine annealing scheduler with a linear warmup. The batch size for training is set at 16,32 \& 64, and the training process spans a total of 200 epochs Fig.\ref{fig:combined-results}. We employ Precision (P), Recall (R), and F1-score (F1) as performance metrics as shown in Fig.\ref{fig:combined-figures-fpr}

\vspace{0.09in}
\vspace{0.09in}

\subsection{Results and Discussion}


\begin{table}[!t]
\caption{Computation Comparison of SOAR with CNN and Transformer based models}
\label{table:computation_marks}
\centering
\begin{tabular}{p{0.12\textwidth} p{0.12\textwidth} p{0.12\textwidth}}
\hline
\textbf{MODEL} & \textbf{PARAMS(M)} & \textbf{GFLOPS} \\ \hline
SNUNet & 10.21 & 176.36 \\ \hline
TransUNetCD & 28.37 & 244.54 \\ \hline
SwinSUNet & 39.28 & 43.50 \\ \hline
SOAR & 17.13 & 45.74 \\ \hline
\end{tabular}
\end{table}

This study has made substantial progress in small object detection in aerial imagery by employing the SAHI framework on YOLO v9 and the Vision Mamba model with a bidirectional State Space Model. These innovations have effectively tackled the challenges of detecting small objects obscured by background noise, enhancing detection accuracy and computational efficiency. The integration of Programmable Gradient Information (PGI) and position embeddings allows for a nuanced, location-aware analysis that is well-suited for remote sensing and computer vision task settings. These findings demonstrate the potential of these lightweight and adaptable models to serve as foundational technologies for future advancements in aerial object recognition and other complex visual tasks. As we continue to refine these methods, they promise to significantly impact the development of computer vision, particularly in resource-limited scenarios.

   
  


\begin{table}[!t] 
\renewcommand{\arraystretch}{1.3} 
\caption{Accuracy Assessment of SOAR with CNN and Transformer Based Models}
\label{table:accuracy} 
\centering 
\begin{tabular}{l c c c c c c } 
\hline
\textbf{MODEL} & \textbf{Rec} & \textbf{Pre} & \textbf{OA} & \textbf{F1} & \textbf{IoU} & \textbf{KC} \\ \hline
SNUNet & 72.21 & 74.09 & 87.49 & 73.14 & 57.66 & 64.99 \\ \hline
TransUNetCD & 77.73 & 82.59 & 90.88 & 80.09 & 66.79 & 74.18 \\ \hline
SwinSUNet & 79.75 & 83.50 & 91.51 & 81.58 & 68.89 & 76.06 \\ \hline
SOAR & 79.59 & 83.06 & 91.36 & 81.29 & 68.48 & 75.68 \\ \hline
\end{tabular}
\end{table}

\vspace{0.1in}
\begin{figure}[H]
    \centering
    \includegraphics[width=0.7\linewidth]{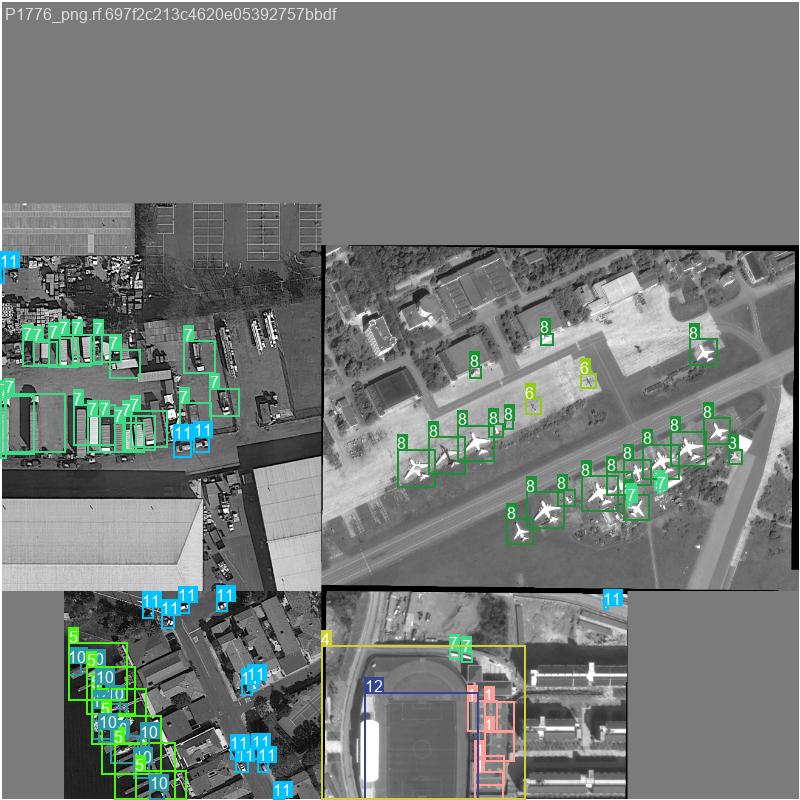}
    \caption{Inferencing Results}
    \label{fig:ir}
\end{figure}
\vspace{0.1in}

\section{Conclusion}
This study has made substantial progress in small object detection in aerial imagery by employing the SAHI framework on YOLO v9 and the Vision Mamba model with a bidirectional State Space Model. These innovations have effectively tackled the challenges of detecting small objects obscured by background noise, enhancing detection accuracy and computational efficiency. The integration of Programmable Gradient Information (PGI) and position embeddings allows for a nuanced, location-aware analysis that is well-suited for remote sensing and computer vision task settings. These findings demonstrate the potential of these lightweight and adaptable models to serve as foundational technologies for future advancements in aerial object recognition and other complex visual tasks. As we continue to refine these methods, they promise to significantly impact the development of computer vision, particularly in resource-limited scenarios.

\vspace{0.2in}

\section{Acknowledgements} The authors express their gratitude to their fellow peers and the High Performace Computing \& Artificial Intelligence Centre at Netaji Subh`as University of Technology for the resources and invaluable feedback during the course of this project.
\\
During the preparation of this work, the authors utilized ChatGPT \cite{openai2023chatgpt}, to assist with generating initial data analysis insights and refining the sections of the manuscript for literature review. After utilizing this tool, the authors carefully reviewed and edited the content as needed and take full responsibility for the content of the publication

\vspace{0.1in}
\begin{figure}[thbp]
    \centering
    \includegraphics[width=0.7\linewidth]{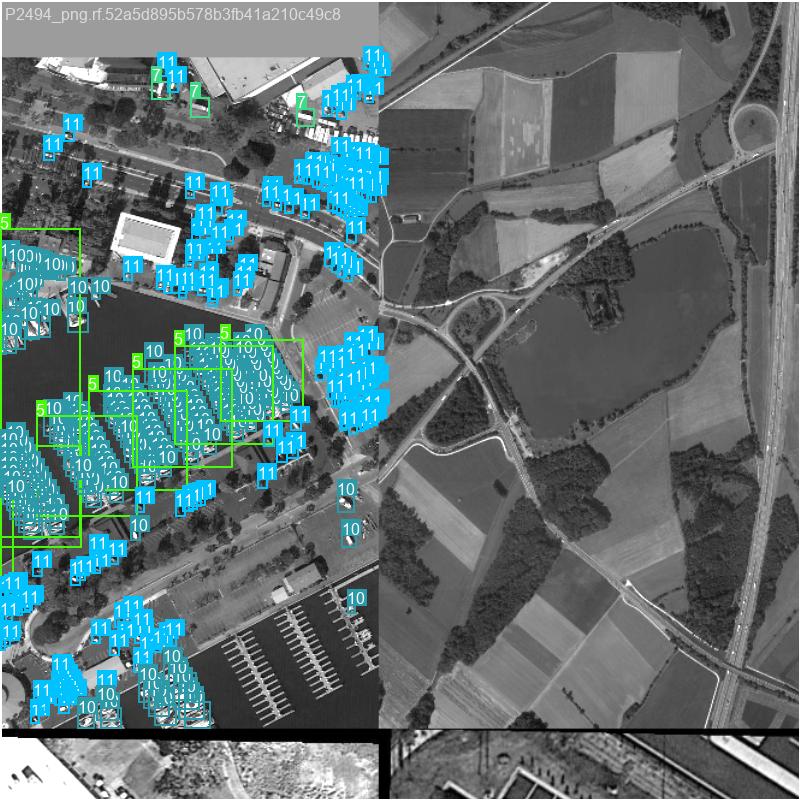}
    \caption{Inferencing Results from another batch}
    \label{fig:ir-2}
\end{figure}
\vspace{0.1in}

\bibliographystyle{IEEEtran}
\bibliography{citation}

\begin{thebibliography}{10}
\providecommand{\url}[1]{#1}
\csname url@samestyle\endcsname
\providecommand{\newblock}{\relax}
\providecommand{\bibinfo}[2]{#2}
\providecommand{\BIBentrySTDinterwordspacing}{\spaceskip=0pt\relax}
\providecommand{\BIBentryALTinterwordstretchfactor}{4}
\providecommand{\BIBentryALTinterwordspacing}{\spaceskip=\fontdimen2\font plus
\BIBentryALTinterwordstretchfactor\fontdimen3\font minus \fontdimen4\font\relax}
\providecommand{\BIBforeignlanguage}[2]{{%
\expandafter\ifx\csname l@#1\endcsname\relax
\typeout{** WARNING: IEEEtran.bst: No hyphenation pattern has been}%
\typeout{** loaded for the language `#1'. Using the pattern for}%
\typeout{** the default language instead.}%
\else
\language=\csname l@#1\endcsname
\fi
#2}}
\providecommand{\BIBdecl}{\relax}
\BIBdecl

\bibitem{lin2014microsoft}
T.-Y. Lin, M.~Maire, S.~Belongie, J.~Hays, P.~Perona, D.~Ramanan, P.~Doll{\'a}r, and C.~L. Zitnick, ``Microsoft coco: Common objects in context,'' in \emph{Computer Vision--ECCV 2014: 13th European Conference, Zurich, Switzerland, September 6-12, 2014, Proceedings, Part V 13}.\hskip 1em plus 0.5em minus 0.4em\relax Springer, 2014, pp. 740--755.

\bibitem{han2021align}
J.~Han, J.~Ding, J.~Li, and G.-S. Xia, ``Align deep features for oriented object detection,'' \emph{IEEE transactions on geoscience and remote sensing}, vol.~60, pp. 1--11, 2021.

\bibitem{yang2021r3det}
X.~Yang, J.~Yan, Z.~Feng, and T.~He, ``R3det: Refined single-stage detector with feature refinement for rotating object,'' in \emph{Proceedings of the AAAI conference on artificial intelligence}, vol.~35, no.~4, 2021, pp. 3163--3171.

\bibitem{yang2019scrdet}
X.~Yang, J.~Yang, J.~Yan, Y.~Zhang, T.~Zhang, Z.~Guo, X.~Sun, and K.~Fu, ``Scrdet: Towards more robust detection for small, cluttered and rotated objects,'' in \emph{Proceedings of the IEEE/CVF international conference on computer vision}, 2019, pp. 8232--8241.

\bibitem{ding2019learning}
J.~Ding, N.~Xue, Y.~Long, G.-S. Xia, and Q.~Lu, ``Learning roi transformer for oriented object detection in aerial images,'' in \emph{Proceedings of the IEEE/CVF conference on computer vision and pattern recognition}, 2019, pp. 2849--2858.

\bibitem{han2021redet}
J.~Han, J.~Ding, N.~Xue, and G.-S. Xia, ``Redet: A rotation-equivariant detector for aerial object detection,'' in \emph{Proceedings of the IEEE/CVF conference on computer vision and pattern recognition}, 2021, pp. 2786--2795.

\bibitem{li2022oriented}
W.~Li, Y.~Chen, K.~Hu, and J.~Zhu, ``Oriented reppoints for aerial object detection,'' in \emph{Proceedings of the IEEE/CVF conference on computer vision and pattern recognition}, 2022, pp. 1829--1838.

\bibitem{ma2018arbitrary}
J.~Ma, W.~Shao, H.~Ye, L.~Wang, H.~Wang, Y.~Zheng, and X.~Xue, ``Arbitrary-oriented scene text detection via rotation proposals,'' \emph{IEEE transactions on multimedia}, vol.~20, no.~11, pp. 3111--3122, 2018.

\bibitem{xu2020gliding}
Y.~Xu, M.~Fu, Q.~Wang, Y.~Wang, K.~Chen, G.-S. Xia, and X.~Bai, ``Gliding vertex on the horizontal bounding box for multi-oriented object detection,'' \emph{IEEE transactions on pattern analysis and machine intelligence}, vol.~43, no.~4, pp. 1452--1459, 2020.

\bibitem{chen2020piou}
Z.~Chen, K.~Chen, W.~Lin, J.~See, H.~Yu, Y.~Ke, and C.~Yang, ``Piou loss: Towards accurate oriented object detection in complex environments,'' in \emph{Computer Vision--ECCV 2020: 16th European Conference, Glasgow, UK, August 23--28, 2020, Proceedings, Part V 16}.\hskip 1em plus 0.5em minus 0.4em\relax Springer, 2020, pp. 195--211.

\bibitem{yang2021rethinking}
X.~Yang, J.~Yan, Q.~Ming, W.~Wang, X.~Zhang, and Q.~Tian, ``Rethinking rotated object detection with gaussian wasserstein distance loss,'' in \emph{International conference on machine learning}.\hskip 1em plus 0.5em minus 0.4em\relax PMLR, 2021, pp. 11\,830--11\,841.

\bibitem{yang2021learning}
X.~Yang, X.~Yang, J.~Yang, Q.~Ming, W.~Wang, Q.~Tian, and J.~Yan, ``Learning high-precision bounding box for rotated object detection via kullback-leibler divergence,'' \emph{Advances in Neural Information Processing Systems}, vol.~34, pp. 18\,381--18\,394, 2021.

\bibitem{lin2017focal}
T.-Y. Lin, P.~Goyal, R.~Girshick, K.~He, and P.~Doll{\'a}r, ``Focal loss for dense object detection,'' in \emph{Proceedings of the IEEE international conference on computer vision}, 2017, pp. 2980--2988.

\bibitem{zhang2021varifocalnet}
H.~Zhang, Y.~Wang, F.~Dayoub, and N.~Sunderhauf, ``Varifocalnet: An iou-aware dense object detector,'' in \emph{Proceedings of the IEEE/CVF conference on computer vision and pattern recognition}, 2021, pp. 8514--8523.

\bibitem{cai2018cascade}
Z.~Cai and N.~Vasconcelos, ``Cascade r-cnn: Delving into high quality object detection,'' in \emph{Proceedings of the IEEE conference on computer vision and pattern recognition}, 2018, pp. 6154--6162.

\bibitem{ren2016faster}
S.~Ren, K.~He, R.~Girshick, and J.~Sun, ``Faster r-cnn: Towards real-time object detection with region proposal networks,'' \emph{IEEE transactions on pattern analysis and machine intelligence}, vol.~39, no.~6, pp. 1137--1149, 2016.

\bibitem{everingham2010pascal}
M.~Everingham, L.~Van~Gool, C.~K. Williams, J.~Winn, and A.~Zisserman, ``The pascal visual object classes (voc) challenge,'' \emph{International journal of computer vision}, vol.~88, pp. 303--338, 2010.

\bibitem{deng2009large}
J.~Deng, ``A large-scale hierarchical image database,'' \emph{Proc. of IEEE Computer Vision and Pattern Recognition, 2009}, 2009.

\bibitem{cagatay2022slicing}
F.~Cagatay~Akyon, S.~Onur~Altinuc, and A.~Temizel, ``Slicing aided hyper inference and fine-tuning for small object detection,'' \emph{arXiv e-prints}, pp. arXiv--2202, 2022.

\bibitem{lin2017feature}
T.-Y. Lin, P.~Doll{\'a}r, R.~Girshick, K.~He, B.~Hariharan, and S.~Belongie, ``Feature pyramid networks for object detection,'' in \emph{Proceedings of the IEEE conference on computer vision and pattern recognition}, 2017, pp. 2117--2125.

\bibitem{zheng2020distance}
Z.~Zheng, P.~Wang, W.~Liu, J.~Li, R.~Ye, and D.~Ren, ``Distance-iou loss: Faster and better learning for bounding box regression,'' in \emph{Proceedings of the AAAI conference on artificial intelligence}, vol.~34, no.~07, 2020, pp. 12\,993--13\,000.

\bibitem{wang2024yolov9}
C.-Y. Wang, I.-H. Yeh, and H.-Y.~M. Liao, ``Yolov9: Learning what you want to learn using programmable gradient information,'' \emph{arXiv preprint arXiv:2402.13616}, 2024.

\bibitem{wang2022designing}
C.-Y. Wang, H.-Y.~M. Liao, and I.-H. Yeh, ``Designing network design strategies through gradient path analysis,'' \emph{arXiv preprint arXiv:2211.04800}, 2022.

\bibitem{wang2020cspnet}
C.-Y. Wang, H.-Y.~M. Liao, Y.-H. Wu, P.-Y. Chen, J.-W. Hsieh, and I.-H. Yeh, ``Cspnet: A new backbone that can enhance learning capability of cnn,'' in \emph{Proceedings of the IEEE/CVF conference on computer vision and pattern recognition workshops}, 2020, pp. 390--391.

\bibitem{akyon2022slicing}
F.~C. Akyon, S.~O. Altinuc, and A.~Temizel, ``Slicing aided hyper inference and fine-tuning for small object detection,'' in \emph{2022 IEEE International Conference on Image Processing (ICIP)}.\hskip 1em plus 0.5em minus 0.4em\relax IEEE, 2022, pp. 966--970.

\bibitem{zhu2024vision}
L.~Zhu, B.~Liao, Q.~Zhang, X.~Wang, W.~Liu, and X.~Wang, ``Vision mamba: Efficient visual representation learning with bidirectional state space model,'' \emph{arXiv preprint arXiv:2401.09417}, 2024.

\bibitem{dosovitskiy2020image}
A.~Dosovitskiy, L.~Beyer, A.~Kolesnikov, D.~Weissenborn, X.~Zhai, T.~Unterthiner, M.~Dehghani, M.~Minderer, G.~Heigold, S.~Gelly \emph{et~al.}, ``An image is worth 16x16 words: Transformers for image recognition at scale,'' \emph{arXiv preprint arXiv:2010.11929}, 2020.

\bibitem{ding2021object}
J.~Ding, N.~Xue, G.-S. Xia, X.~Bai, W.~Yang, M.~Y. Yang, S.~Belongie, J.~Luo, M.~Datcu, M.~Pelillo \emph{et~al.}, ``Object detection in aerial images: A large-scale benchmark and challenges,'' \emph{IEEE transactions on pattern analysis and machine intelligence}, vol.~44, no.~11, pp. 7778--7796, 2021.

\bibitem{openai2023chatgpt}
OpenAI, ``Chatgpt,'' \url{https://www.openai.com/chatgpt}, 2023, accessed: [insert date of access].

\end{thebibliography}


\end{document}